\documentclass{article}
\usepackage{spconf,amsmath,graphicx}
\usepackage{verbatim}
\usepackage{amssymb}

\title{DOES AN ENSEMBLE OF GANS LEAD TO BETTER PERFORMANCE WHEN TRAINING SEGMENTATION NETWORKS WITH SYNTHETIC IMAGES?}
\name{Måns Larsson$^{\: a}$, Muhammad Usman Akbar$^{\: b,d}$, Anders Eklund$^{\: b,c,d}$}

\address{$^a$ Eigenvision, Malmö, Sweden \\ $^b$ Division of Medical Informatics, Department of Biomedical Engineering\\ $^c$ Division of Statistics \& Machine Learning, Department of Computer and Information Science\\ $^d$ Center for Medical Image Science and Visualization (CMIV) \\Link\"{o}ping University, Sweden \\ }
\begin{document}
%
\maketitle

\begin{abstract}
Large annotated datasets are required to train segmentation networks. In medical imaging, it is often difficult, time consuming and expensive to create such datasets, and it may also be difficult to share these datasets with other researchers. Different AI models can today generate very realistic synthetic images, which can potentially be openly shared as they do not belong to specific persons. However, recent work has shown that using synthetic images for training deep networks often leads to worse performance compared to using real images. Here we demonstrate that using synthetic images and annotations from an ensemble of 20 GANs, instead of from a single GAN, increases the Dice score on real test images with 4.7\% to 14.0\% on specific classes.
\end{abstract}
\vspace{-0.1cm}
\section{Introduction}
Artificial intelligence (AI) has become a popular tool in medicine to help clinicians work more efficiently and has been vastly discussed in recent decades~\cite{jiang2017artificial}. Due to the recent success of machine learning (ML) methods in other domains and the steady increase of image-based medical examinations, applying ML methods for medical image analysis is becoming increasingly popular. Since ML models learn from data, the quantity and quality of medical image data have a direct impact on the success of ML-based algorithms. These algorithms require large datasets with annotations created by human experts. Unfortunately, the availability of large open-source datasets in biomedical imaging is low. The main reasons for the unavailability of large public datasets in medical imaging can be categorized into three main groups; privacy concerns (due to the potentially sensitive information of the patients), expensiveness in terms of cost and time for data collection and annotation, and lack of data due to the rarity of some diseases. Synthetic images overcome all these problems and can help researchers in training complex models. Recently, augmentation of datasets with synthetically generated images is one of the techniques used to explore the insufficiency of annotated data to train the ML algorithms~\cite{shin2018medical}. These synthetically produced images can be used for training deep networks, and can potentially be shared among peers without breaking privacy laws and data protection regulations. 

Generative adversarial networks (GANs) recently gained a lot of attention in the computer vision community. Generative modeling is an unsupervised machine learning task that involves discovering and learning the patterns automatically in the input data, and then training itself to further generate new examples. 
GANs have proven themselves to be useful in multiple scenarios such as data augmentation~\cite{cha2019evaluation}, image-to-image translation~\cite{kang2021synthetic}, domain adaption, and filling in missing information. They provide an opportunity to produce large annotated medical datasets by producing realistic synthetic images which can further be used to train deep models. There are also several examples where GANs have been succesfully used for medical image applications~\cite{frid2018gan, madani2018chest, bowles2018gan}. \\
GANs work by learning a complex high-dimensional data distribution to generate diverse high-resolution images. For image segmentation, it is necessary to produce annotations for each synthetic image, and different approaches have been used to achieve this. Bowles et al.~\cite{bowles2018gan} used a two-channel approach to synthesize images with their corresponding annotations. Foroozandeh1 et al.~\cite{foroozandeh} used a two-step process where the authors first produce the annotations using a noise-to-image GAN~\cite{karras}, and then used an image-to-image GAN~\cite{park2019semantic} to synthesize MR images from the annotations previously generated. Guibas et al.~\cite{guibas2017synthetic} also used a similar two-step approach to produce images along with corresponding annotations.  

While the main principle of training networks with synthetic images has been demonstrated, several papers have shown that using the synthetic images results in lower performance when training a CNN~\cite{shin2018medical,cha2019evaluation,foroozandeh,eilertsen}.
Eilertsen et al.~\cite{eilertsen} demonstrated that using an ensemble of GANs leads to higher accuracy when training classifiers with synthetic images. Therefore we investigate if using an ensemble of GANs leads to better performance when training segmentation networks with synthetic images and annotations.
In this paper, we follow the multi-channel approach to produce images and their corresponding annotations at the same time.

\vspace{-0.5cm}
\section{Data}
\vspace{-0.1cm}
The MR images used for this project were downloaded from the Multimodal Brain tumour Segmentation Challenge (BraTS) 2020~\cite{bakas3,bakas4,bakas1,bakas2,menze}. The training set contains MR volumes of shape 240 × 240 × 155 from 369 patients, and for each patient four types of MR images are available: T1-weighted (T1), post gadolinium contrast T1-weighted (T1Gd), T2-weighted (T2), and T2 fluid attenuated inversion recovery (FLAIR). The annotations cover three parts of the brain tumor: peritumoural edema (ED), necrotic and non-enhancing tumour core (NCR/NET), and GD-enhancing tumour (ET). We used 313 subjects for training and 56 subjects for testing.

\vspace{-0.35cm}
\section{Methods}

All 3D volumes were split into 2D slices, as a 2D GAN was used, and only slices with at least 15\% pixels with an intensity of more than 50 were included in the training. This resulted in a total of 23,478 5-channel images from the 313 training subjects, and 4,238 5-channel images from the 56 test subjects. Each slice was zero padded from 240 x 240 to 256 x 256 pixels, as the used GAN only works for resolutions that are a power of 2, and the intensity was rescaled to 0 - 255. The intensities for the tumor annotations were changed from [1,2,4] to [51,102,204], such that the intensity range is more similar for the 5 channels.

\vspace{-0.2cm}
\subsection{GAN training}

The progressive growing GAN architecture~\cite{karras} was used for all experiments, with the default settings. The code was modified to generate 5-channel images (T1, T1Gd, T2, FLAIR, segmentation annotations) instead of 3-channel color images. The GAN will thereby learn to jointly generate 4-channel MR images and the corresponding tumor annotations at the same time.

The same GAN was trained 20 times and due to several sources of randomness, each GAN will learn to sample from slightly different parts of the high dimensional image distribution. The total training time was about 100 days on an Nvidia V100 GPU. A total of 100,000 synthetic 5-channel images were generated from each trained GAN. When using images from 5 GANs, each GAN provided 20,000 synthetic images. When using images from 10 GANs, each GAN provided 10,000 synthetic images. 

There is no guarantee that the synthetic annotations will be restricted to the same values as the real annotations ([51,102,204]). The synthetic annotations were therefore thresholded to the closest original annotation value.

\subsection{Tumor segmentation}

The synthetic images were used to train a model to perform tumor segmentation, and the evaluation was performed using real images in the test set. The model structure and training setup used was inspired by the 2D segmentation code from nnUNet~\cite{isensee2019automated}. The model was a U-Net~\cite{ronneberger} with an extra depth layer and instance normalization instead of batch normalization. In addition, the ReLU activations was swapped for leaky ReLUs with a negative slope of $10^{-2}$. All models were trained with a loss consisting of a cross-entropy term and a soft Dice term weighted equally. In addition, deep supervision was used, meaning that the loss was applied on the five highest depth level with weighting $0.5^d$ where $d$ is the depth.

The loss was minimized using stochastic gradient descent with Nesterov momentum of $0.99$ and weight decay of $3 \cdot 10^{-5}$. The initial learning rate was $5 \cdot 10^{-2}$ and was decreased using polynomial learning rate decay with an exponent of $0.9$. All models were trained for $3 \cdot 10^7$ samples or 3 days, whichever occured first. 20\% of the available images were used for validation and the model with the best mean Dice over the validation set was used for evaluation. If both real and synthetic data was used during training, the real dataset and the synthetic dataset were sampled equally often.

During training a geometric, and intensity-augmentation was applied. The image and target is first randomly rotated and scaled. Both rotation and scaling is applied with a probability of 0.75, the image and target is rotated with an angle uniformly sampled from $[-30^{\circ}, 30^{\circ}]$ and the width and height is scaled (independently) with a scale factor uniformly sampled from $[0.9, 1.1]$. Then the four input channels are augmented by; adding Gaussian noise (applied with probability 0.5, zero mean and standard deviation uniformly sampled from $[0.0, 0.05]$), blurring the image (applied with probability 0.2, gaussian blurring with standard deviation sampled from $[0.5, 1.0]$), faking lower resolution (zooming with a factor between 0.75 and 1.0 and then upsampling) and changing the gamma factor (scaling it with a factor between 0.8 and 1.2). Lastly the input image channels are normalized using Z-score normalization.

When training the 20 different GANS, the random training is a positive side effect. However, when training the segmentation networks this effect is undesired. Each segmentation network was therefore trained 10 times.
\vspace{-0.1cm}
\section{Results}
\vspace{-0.1cm}
Figure 1 shows an example of a real 5-channel image and a randomly selected synthetic 5-channel image. 

\begin{figure*}[htb]

\begin{minipage}[b]{1.0\linewidth}
  \centering
  \includegraphics[width=3.1cm]{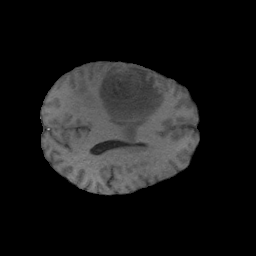}
  \includegraphics[width=3.1cm]{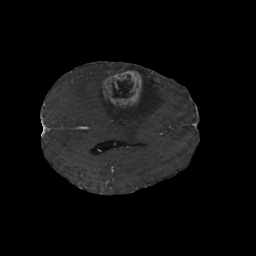}
  \includegraphics[width=3.1cm]{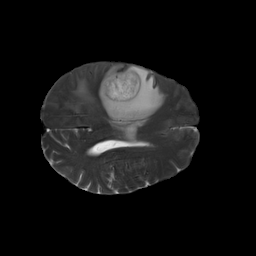}
  \includegraphics[width=3.1cm]{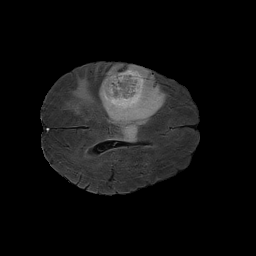}
  \includegraphics[width=3.1cm]{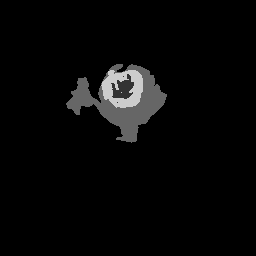}

  \includegraphics[width=3.1cm]{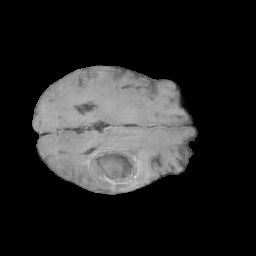}
  \includegraphics[width=3.1cm]{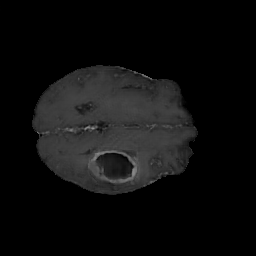}
  \includegraphics[width=3.1cm]{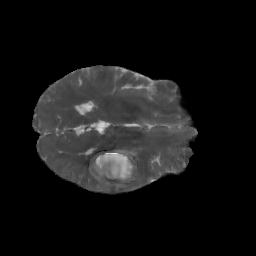}
  \includegraphics[width=3.1cm]{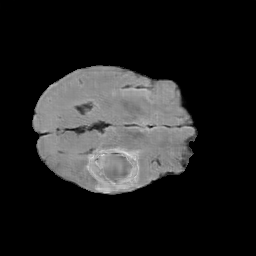}
  \includegraphics[width=3.1cm]{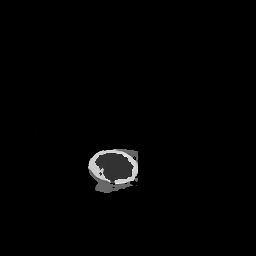}
  
  \caption{Top: a real 5-channel image. Bottom: a synthetic 5-channel image. From left to right: T1-weighted, T1-weighted after gadolinium contrast, T2-weighted, FLAIR, segmentation mask.}
  
\end{minipage}

\end{figure*}

\begin{table*} \footnotesize
    \centering
    \begin{tabular}{c|c|ccc|c}
        \textbf{Original data} & \textbf{\# GANs} & \textbf{ET} & \textbf{ED} & \textbf{NCR/NET} & \textbf{Mean} \\ \hline
\checkmark & 0 & $0.791 \pm 0.009$ & $0.785 \pm 0.003$ & $0.610 \pm 0.008$ & $0.729 \pm 0.004$ \\
\checkmark & 1 & $0.789 \pm 0.010$ & $0.790 \pm 0.008$ & $0.609 \pm 0.013$ & $0.730 \pm 0.008$ \\
\checkmark & 5 & $0.795 \pm 0.012$ & $0.791 \pm 0.004$ & $0.617 \pm 0.008$ & $0.734 \pm 0.006$ \\
\checkmark & 10 & $0.791 \pm 0.008$ & $0.794 \pm 0.005$ & $0.620 \pm 0.009$ & $0.735 \pm 0.003$ \\
\checkmark & 20 & $0.799 \pm 0.008$ & $0.791 \pm 0.005$ & $0.613 \pm 0.011$ & $0.734 \pm 0.006$ \\
 & 1 & $0.665 \pm 0.011$ & $0.632 \pm 0.027$ & $0.478 \pm 0.019$ & $0.592 \pm 0.016$ \\
 & 5 & $0.689 \pm 0.012$ & $0.665 \pm 0.023$ & $0.529 \pm 0.015$ & $0.628 \pm 0.014$ \\
 & 10 & $0.696 \pm 0.012$ & $0.702 \pm 0.013$ & $0.545 \pm 0.012$ & $0.648 \pm 0.008$ \\
 & 20 & $0.696 \pm 0.017$ & $0.693 \pm 0.021$ & $0.555 \pm 0.013$ & $0.648 \pm 0.014$ \\
  
    \end{tabular}
    \caption{Results on the test dataset (56 subjects), Dice calculated per subject. Mean and standard deviation of Dice score across the dataset are calculated for the labels; GD-enhancing tumor (ET), peritumoral edema (ED), and necrotic and non-enhancing tumor core (NCR/NET). All results are presented as mean $\pm$ standard deviation of 10 trainings.}
    \label{tab:res_ens}
\end{table*}

Table~\ref{tab:res_ens} shows the segmentation accuracy on real images when training the U-Net using only real images, using real and synthetic images together, or only using synthetic images. The results are repeated for synthetic images from a single GAN, from 5 GANs, from 10 GANs and 20 GANs. The results represent the mean and standard deviation over 10 trainings of the U-Net. The mean Dice score increases 4.7\% for the enhancing tumor, 11.1\% for the edema part, and 14.0\% for the necrotic core, when training with only synthetic images from 10 GANs instead of from a single GAN. As expected, adding synthetic data to the original dataset during training (row 1-3 in~Table~\ref{tab:res_ens}) only gives a very small increase in segmentation performance. Since the GANs that generate the synthetic data are trained on the same dataset, no new information in the form of annotated data is added and the addition of synthetic data can instead be compared to advanced data augmentation. However, the same trend with more GANs giving better performance still holds for these experiments. For 5 GANs, the improvement compared to 1 GAN was significant (p $<$ 0.05) for NCR/NET and ET. For 10 GANs, the improvement compared to 1 GAN was significant for all three tumor types. Using 10 GANs was significantly better than 5 GANs for ED and ET. The p-values were obtained through a non-parametric sign flipping test, and multiple comparison correction was done using Bonferroni.

Figure 2 shows segmentation results for a randomly selected slice, for the same training setups as in Table 1.

\begin{figure*}
\centering
    \begin{tabular}{ccccc}
        annotation & 1 GAN & 5 GANs & 10 GANs & 20 GANs \\
        \includegraphics[width=0.16\textwidth]{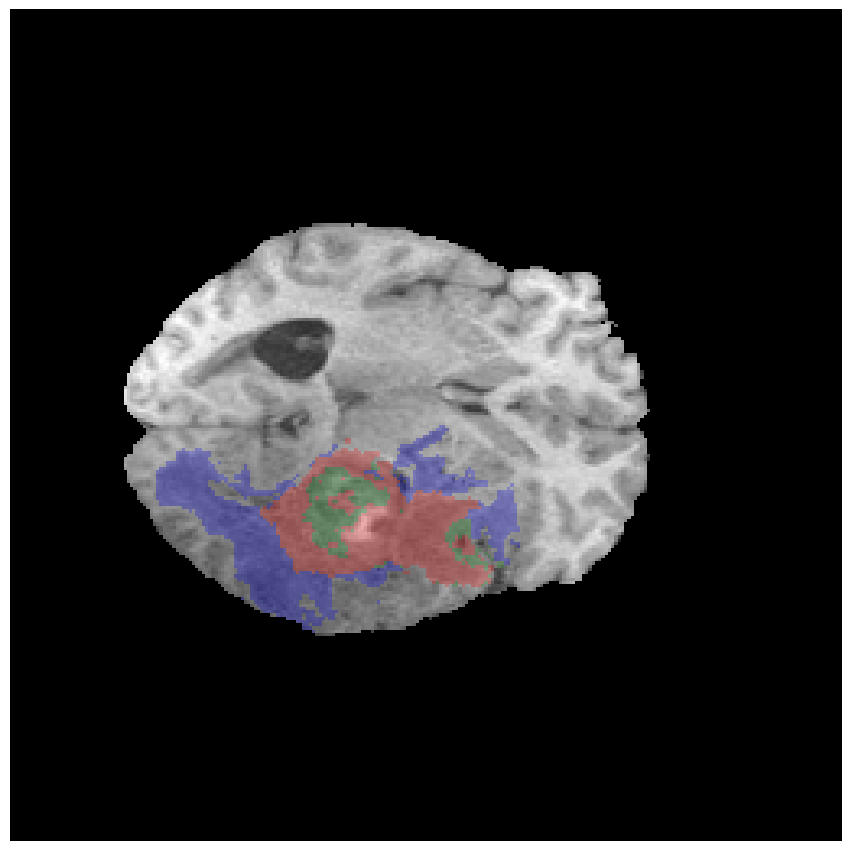} &
        \includegraphics[width=0.16\textwidth]{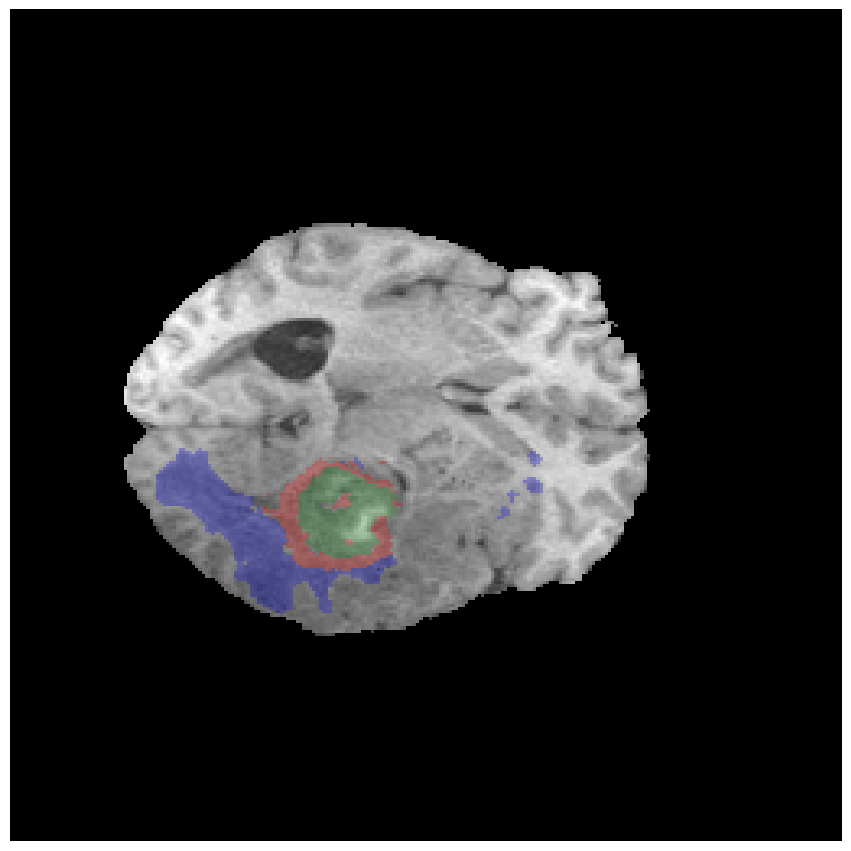} &
        \includegraphics[width=0.16\textwidth]{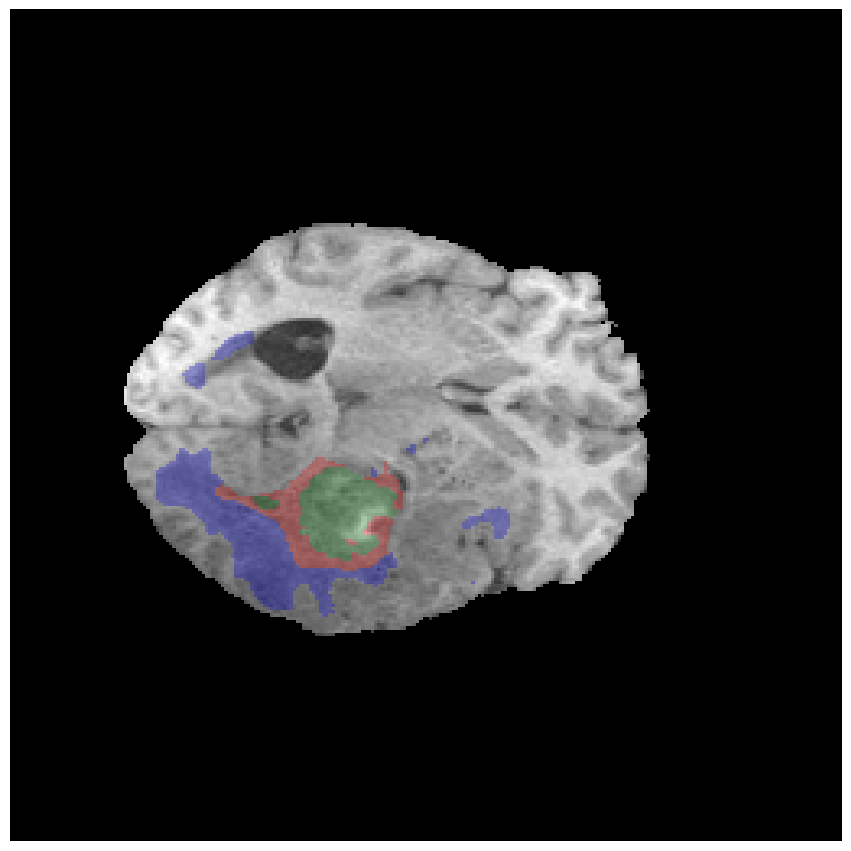} &
        \includegraphics[width=0.16\textwidth]{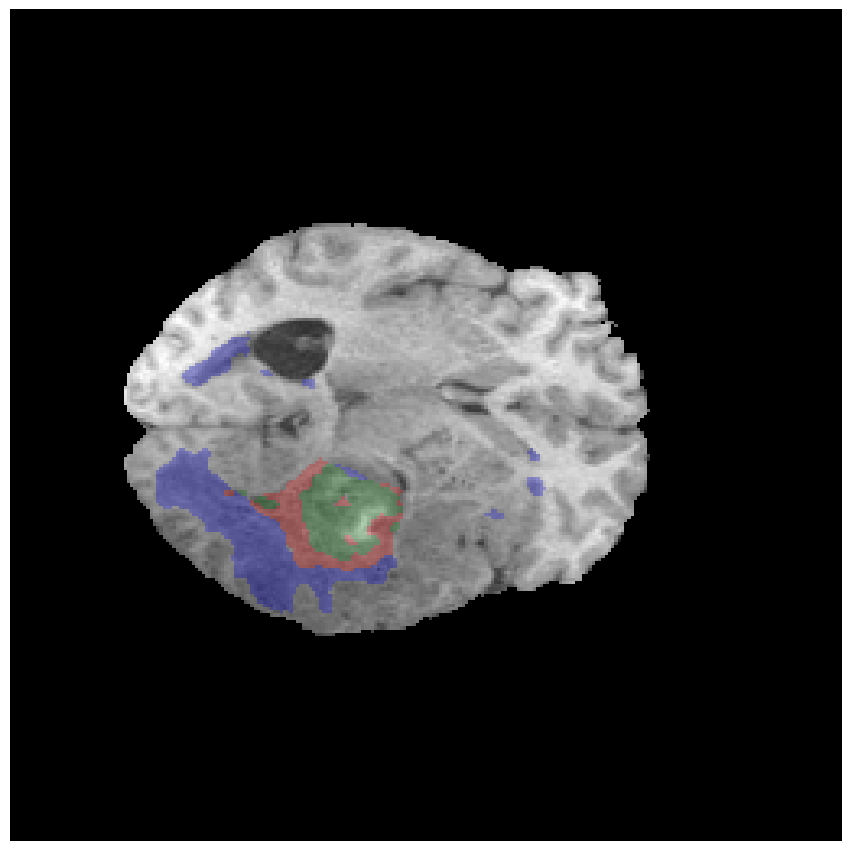} &
        \includegraphics[width=0.16\textwidth]{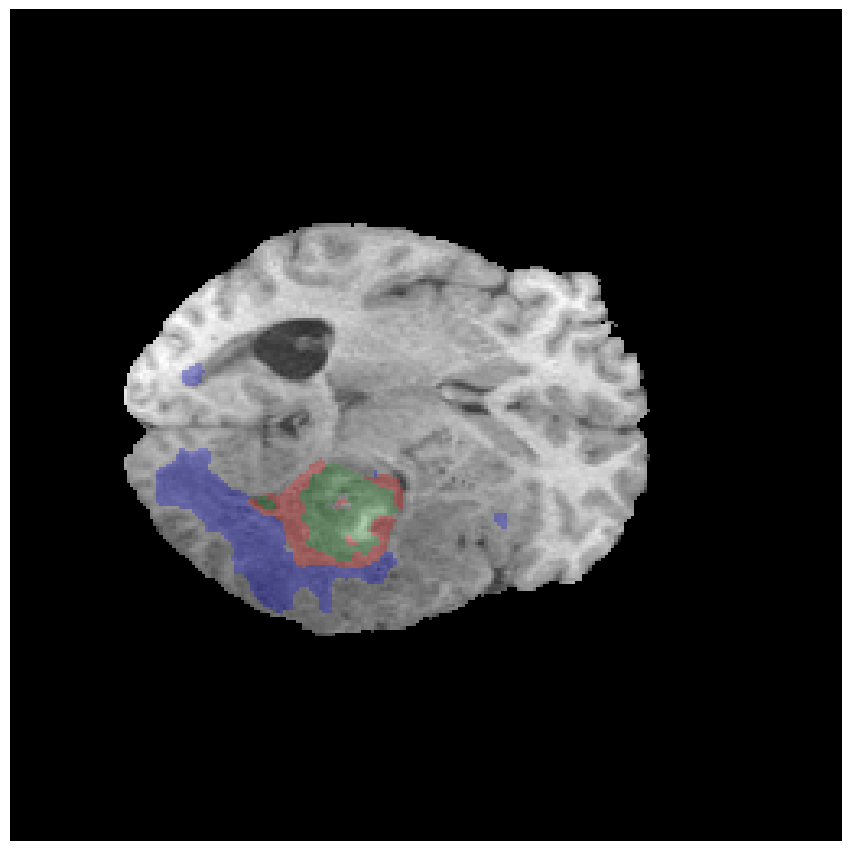} \\
        original & original + 1 GAN & original + 5 GANs & original + 10 GANs & original + 20 GANs \\
         \includegraphics[width=0.16\textwidth]{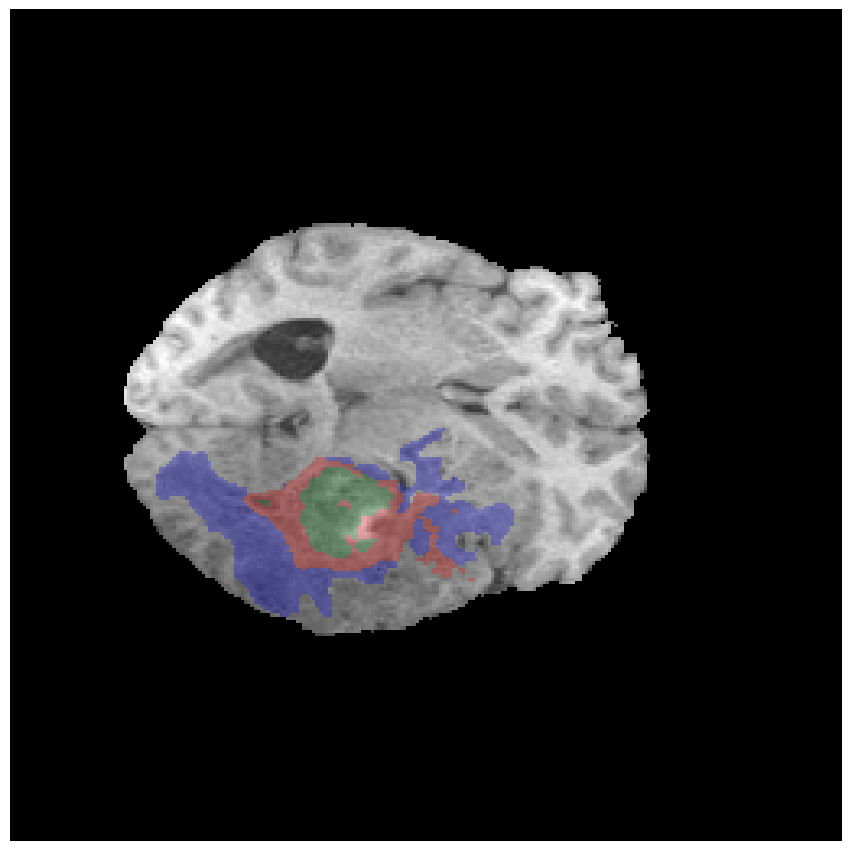} &
        \includegraphics[width=0.16\textwidth]{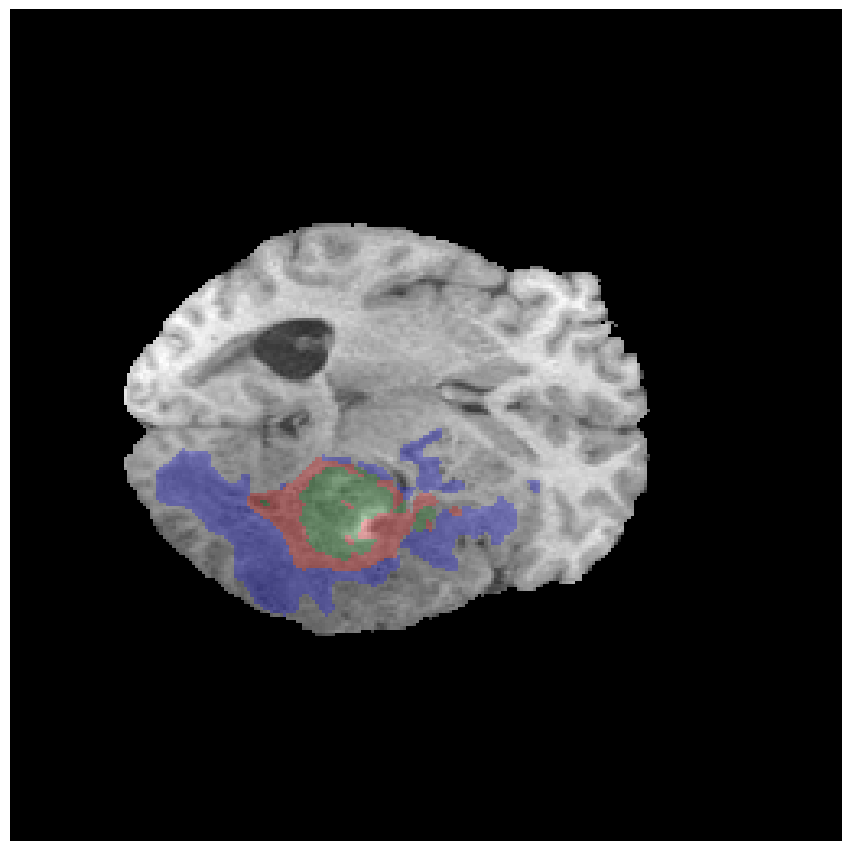} &
        \includegraphics[width=0.16\textwidth]{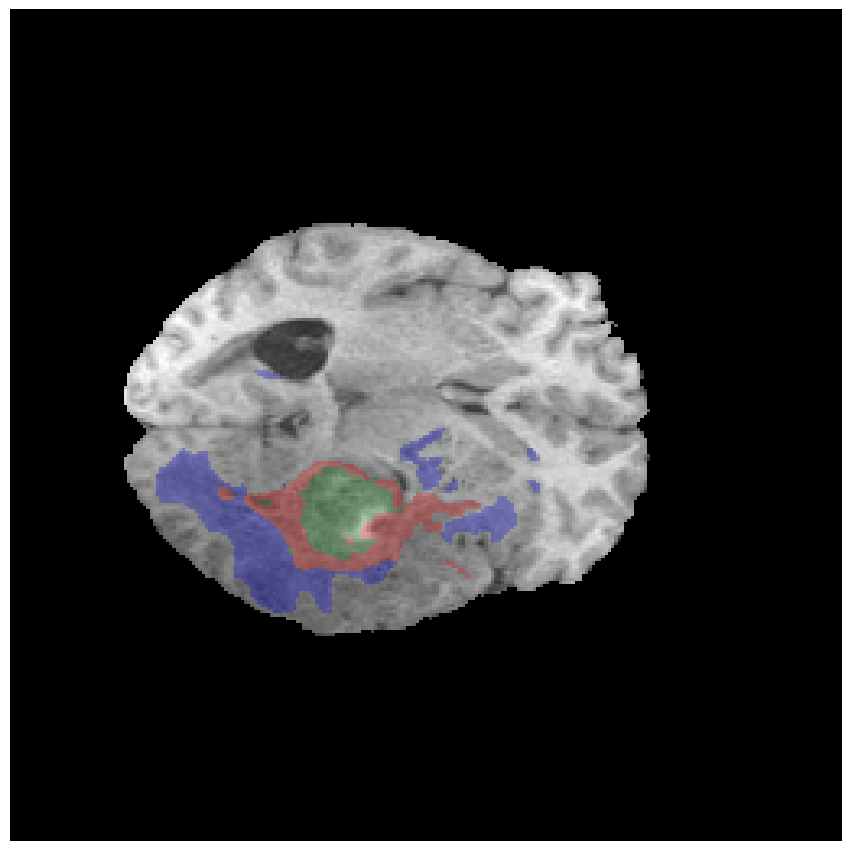} &
        \includegraphics[width=0.16\textwidth]{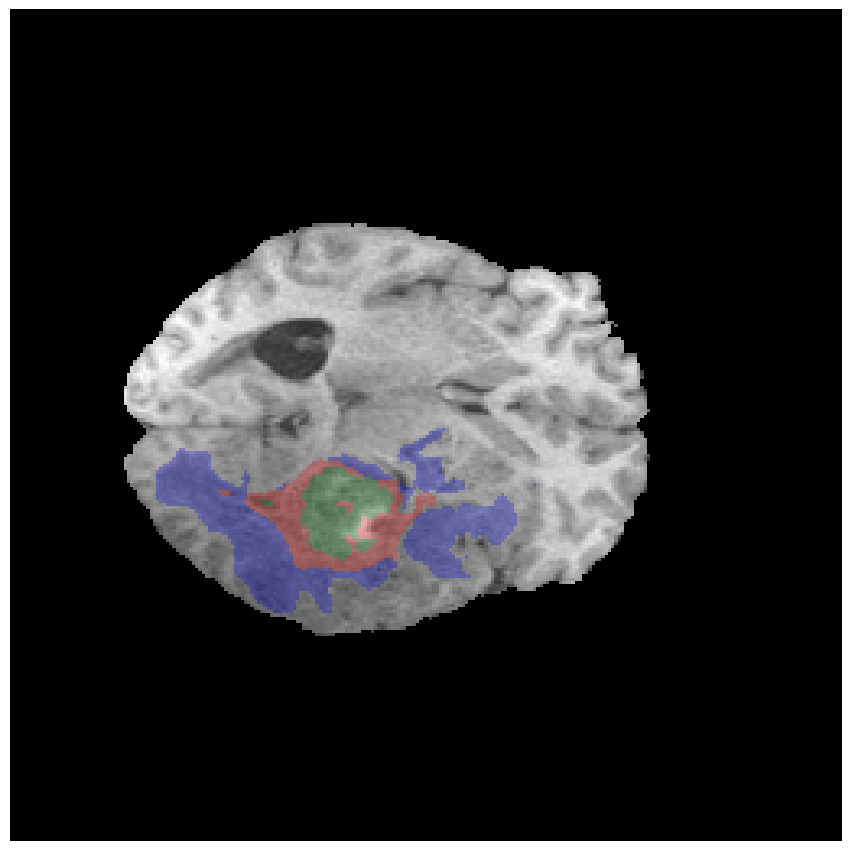} &
        \includegraphics[width=0.16\textwidth]{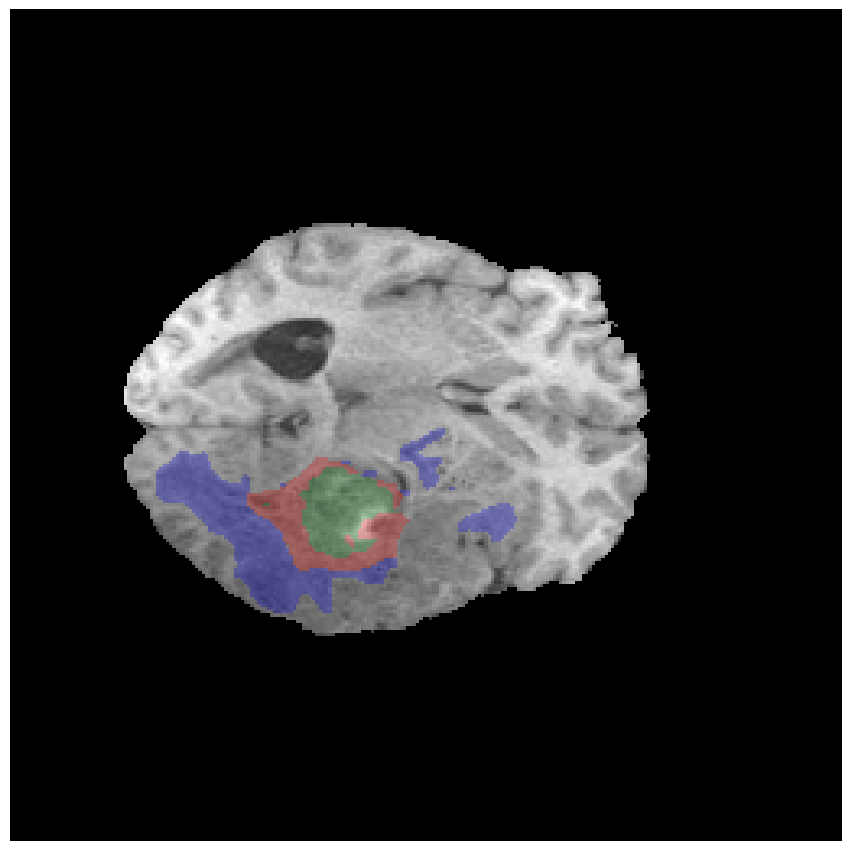} \\
    \end{tabular}
    \caption{Example predictions on an image in the test set. Classes are visualized as colored overlay where red is GD-enhancing tumor, blue is peritumoral edema (ED) and green is necrotic and non-enhancing tumor core (NCR/NET).}
\end{figure*}
\vspace{-0.2cm}
\section{Discussion}
\vspace{-0.2cm}
Our results show that an ensemble of GANs indeed improves the segmentation accuracy when using the synthetic images for training segmentation networks, compared to using synthetic images from a single GAN. This comes at the cost of an increased training time, which scales linearly with the number of GANs. Our results thereby reflect the results obtained by~\cite{eilertsen} for classification. Better results were not obtained when using 20 instead of 10 GANs. Here we use a rather strong baseline model, which may explain why we see only minor improvements when adding synthetic images to the real ones.

A challenge with the BraTS dataset is that the data originates from 19 different centers, and that the image quality and appearance varies between the centers. It is possible that better results can be obtained if the site information is used as a variable in the GAN training, to train conditional GANs which can learn to generate images that represent each center. Although BraTS 2020 contains images from 369 subjects (training set), it is possible that instead using BraTS 2021 (which contains 1251 training subjects) will give substantially better results.

Training with synthetic images still results in substantially lower segmentation accuracy, compared to training with real images. In future work we therefore plan to repeat the experiments with the more recent architectures StyleGAN~\cite{karras2019style}, StyleGAN2~\cite{karras2020analyzing} and StyleGAN3~\cite{karras2021alias} (although they are even more computationally demanding), as well as with more recent diffusion models. It would especially be interesting to train conditional models where the imaging site, tumor size and tumor location can be controlled. We also plan to look into 3D GANs~\cite{eklund} and 3D diffusion models~\cite{pinaya}, but a general problem is that they will use a lot of GPU memory for generating 5-channel volumes.

\vspace{-0.2cm}
\section{Compliance with ethical standards}
\vspace{-0.2cm}
This research study was conducted retrospectively using human subject data made available as open access by BraTS. Ethical approval was not required as confirmed by the ethical review board of Linköping.

\vspace{-0.2cm}
\section{Acknowledgements}
\vspace{-0.2cm}
Training the segmentation networks was performed using the supercomputing resource Berzelius (480 Nvidia A100 GPUs) provided by National Supercomputer Centre at Linköping University, Sweden, and the Knut and Alice Wallenberg Foundation. 

This research was supported by the ITEA/VINNOVA project ASSIST (Automation, Surgery Support and Intuitive 3D visualization to optimize workflow in IGT SysTems, 2021-01954), the VINNOVA funded project AIDA, LiU Cancer and the Åke Wiberg foundation. Anders Eklund has previously received graphics hardware from Nvidia.

\bibliographystyle{IEEEbib}
\bibliography{strings,refs}

\end{document}